\documentclass[10pt,twocolumn,letterpaper]{article}

\usepackage{cvpr}              % To produce the CAMERA-READY version

\usepackage{algorithm}
\usepackage{algorithmic}
\usepackage{amsthm,amsmath,amsfonts,amssymb,bm,xspace,nicefrac}
\usepackage{cases}

\usepackage{microtype}
\usepackage{cite}

\usepackage{xcolor}
\usepackage{color}
\usepackage{colortbl}

\usepackage{enumitem}
\usepackage{booktabs}
\usepackage{multirow}
\usepackage{makecell}
\usepackage{array}
\usepackage{tabulary}
\usepackage{threeparttable}
\usepackage[switch]{lineno}
\usepackage{hhline}
\usepackage{tabu}
\usepackage{boldline}

\usepackage{graphicx}
\usepackage{amsmath}
\usepackage{amssymb}
\usepackage{booktabs}

\usepackage{caption}

\usepackage[pagebackref,breaklinks,colorlinks]{hyperref}

\newcommand{\argmax}{\mathop{\arg\max}}

\usepackage{color}

\usepackage[capitalize]{cleveref}
\crefname{section}{Sec.}{Secs.}
\Crefname{section}{Section}{Sections}
\Crefname{table}{Table}{Tables}
\crefname{table}{Tab.}{Tabs.}

\def\cvprPaperID{2016} % *** Enter the CVPR Paper ID here
\def\confName{CVPR}
\def\confYear{2022}

\begin{document}

\title{Meta-attention for ViT-backed Continual Learning}
\makeatletter
\newcommand{\myfnsymbol}[1]{%
  \expandafter\@myfnsymbol\csname c@#1\endcsname
}
\newcommand{\@myfnsymbol}[1]{%
  \ifcase #1
    % 0
  \or \TextOrMath{\textdagger}{\dagger}% 2
  \fi
}

\author {%
    % Authors
    Mengqi Xue\textsuperscript{$1$},
    Haofei Zhang\textsuperscript{$1$},
    Jie Song\textsuperscript{$1,\dagger$},
    Mingli Song\textsuperscript{$1,2$} \\
    \textsuperscript{$1$}Zhejiang University \\
    \textsuperscript{$2$}Alibaba-Zhejiang University Joint Research Institute of Frontier Technologies, Zhejiang University \\
     {\tt\small \{mqxue,haofeizhang,sjie,brooksong\}@zju.edu.cn}\\

}

% Thanks notes for title uses \myfnsymbol
\renewcommand{\thefootnote}{\myfnsymbol{footnote}}
\maketitle
\footnotetext[1]{Corresponding author}%
\setcounter{footnote}{0}% Restart footnote counter
% Footnotes for rest of document uses \fnsymbol (or whatever you choose)
\renewcommand{\thefootnote}{\arabic{footnote}}
% \author{Mengqi Xue\\
% Zhejiang University\\
% {\tt\small mqxue@zju.edu.cn }
% \and
\maketitle

%%%%%%%%% ABSTRACT
\begin{abstract}

Continual learning is a longstanding research topic due to its crucial role in tackling continually arriving tasks. Up to now, the study of continual learning in computer vision is mainly restricted to convolutional neural networks~(CNNs). However, recently there is a tendency that the newly emerging vision transformers~(ViTs) are gradually dominating the field of computer vision, which leaves CNN-based continual learning lagging behind as they can suffer from severe performance degradation if straightforwardly applied to ViTs. In this paper, we study ViT-backed continual learning to strive for higher performance riding on recent advances of ViTs. Inspired by mask-based continual learning methods in CNNs, where a mask is learned per task to adapt the pre-trained ViT to the new task, we propose \textbf{MEta-ATtention} (MEAT), i.e., attention to self-attention, to adapt a pre-trained ViT to new tasks without sacrificing performance on already learned tasks. Unlike prior mask-based methods like Piggyback, where all parameters are associated with corresponding masks, MEAT leverages the characteristics of ViTs and only masks a portion of its parameters. It renders MEAT more efficient and effective with less overhead and higher accuracy. Extensive experiments demonstrate that MEAT exhibits significant superiority to its state-of-the-art CNN counterparts, with $4.0\sim 6.0\%$ absolute boosts in accuracy. Our code has been released at\textit{~\url{https://github.com/zju-vipa/MEAT-TIL}}.

\end{abstract}

%%%%%%%%% BODY TEXT
\section{Introduction}
\label{intro}
Being capable of tackling everchanging tasks is a favorable merit in open-world scenarios. Humans excel at solving constantly emerging tasks by associating them with previously learned knowledge. Deep neural networks~(DNNs), however, usually suffer from \textit{catastrophic forgetting}~\cite{mccloskey1989catastrophic} if simply adapted to new tasks due to the differences between tasks in data biases.

\begin{figure}[!t]
\centering%
\includegraphics[width=1.0\linewidth]{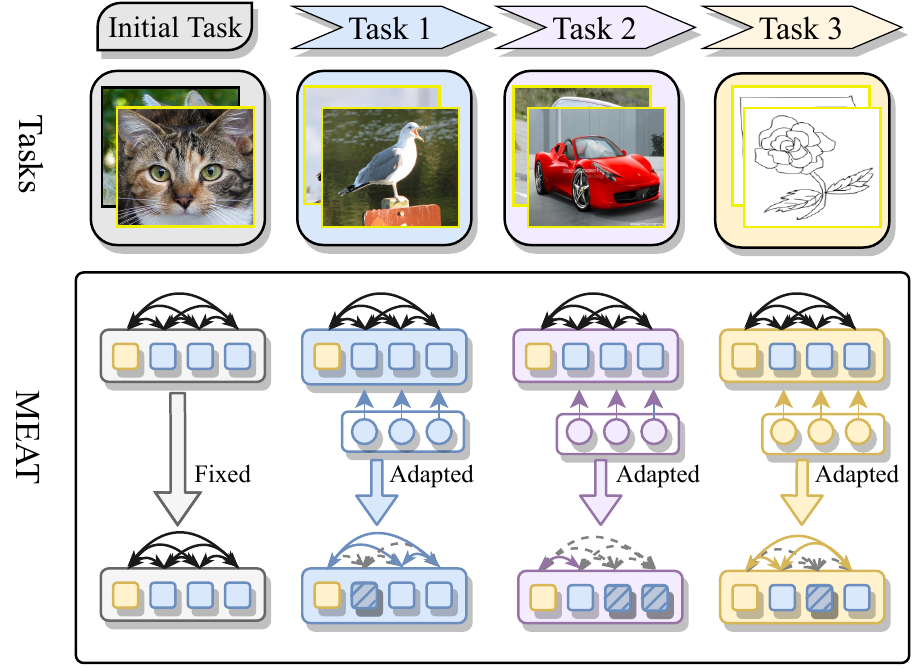}

\caption{The proposed MEAT for task continual learning in the MHSA block with vision transformers. With the increase of new tasks, MEAT dynamically assigns attention masks to generate task-specific self-attention patterns per task.}
\label{fig:incremental}
\end{figure}

Over the past years, large pieces of literature have been devoted to addressing the catastrophic forgetting problem to enable DNNs to master new-arrived tasks in a sequence~\cite{rusu2016progressive, li2017learning, rebuffi2017icarl, kirkpatrick2017overcoming, mallya2018piggyback, singh2020calibrating, serra2018overcoming}. Existing continual learning methods can be broadly categorized into three schools: \textit{replay methods}~\cite{rebuffi2017icarl, chaudhry2019continual, rolnick2018experience, isele2018selective, de2021CoPE}, \textit{regularization methods}~\cite{li2017learning, jung2016less, kirkpatrick2017overcoming, zenke2017continual, zhang2020class, lee2017imm, paik2020npc} and \textit{mask methods}~\cite{mallya2018packnet, mallya2018piggyback, masana2021ternary, serra2018overcoming, hung2019compacting}. Replay methods replay previous task samples, which are stored in raw format or generated with a generative model, to alleviate forgetting while learning a new task. To avoid storing raw inputs, prioritize privacy and alleviate memory requirements, regularization methods introduce a regularization term to consolidate previous knowledge 
while learning the new task. Mask methods learn a mask per task to adapt the pre-trained model to the new task for preventing any possible forgetting.

Albeit remarkable progress made in computer vision, most of the aforementioned methods are tailored for CNNs for their dominant performance in the field over the past decade. However, the primacy of CNNs in computer vision is recently challenged by vision transformers~(ViTs)~\cite{dosovitskiy2020image, liu2021swin, touvron2021training}, due to the more general-purpose architecture (\ie, bridging the architecture gap between natural language processing and computer vision) and superior performance of ViTs. In contrast to the rapid development of ViTs, prior CNN-based continual learning methods appear a bit outdated as straightforwardly applying them to ViTs does not take full advantage of the characteristics of transformers. 

In this work, we devote ourselves to ViT-backed continual learning to keep pace with the advancement of ViTs. Specifically, we ground our proposed method on mask method~\cite{mallya2018packnet, mallya2018piggyback, serra2018overcoming, masana2021ternary} for the following three main reasons: (1) mask methods in fact dedicate different parameters to each task, thus perfectly bypassing the catastrophic forgetting problem; (2) mask methods are not sensitive to task order, which is a very favorable merit in continual learning; (3) mask methods avoid expensive data storage and exhibit a larger capacity to handle more tasks, which gives them a prominent edge over replay and regularization methods. Motivated by these attractive advantages, we propose our ViT-backed mask-based continual learning method, dubbed as \textit{MEta-ATtention}~(MEAT), to further boost the continual learning performance, as illustrated in Figure~\ref{fig:incremental}. MEAT inherits all the aforementioned merits, and meanwhile introduces the following innovations that distinguish it from prior mask methods: (1) MEAT fully leverages the architectural characteristics of ViTs and introduces the \textit{attention to self-attention} (where the meta-attention comes) mechanism, which is tailored for transformer-based architectures and makes MEAT furthermore effective.
(2) Prior methods, like Piggyback~\cite{mallya2018piggyback}, require manually setting the threshold hyper-parameter for binarizing the mask. MEAT adopts Gumbel-softmax trick~\cite{jang2016categorical} to resolve the optimization difficulty of discrete mask values, which relaxes the burden of the hyper-parameter 
search. 
(3) Unlike prior mask-based methods where all parameters are assigned to masks, MEAT introduces masks to only a portion of its parameters, which renders it more efficient than prior methods.

To validate the superiority of the proposed method, extensive experiments, including benchmark comparison and ablation study, are conducted on a diverse set of image classification benchmarks (including ImageNet~\cite{deng2009imagenet}, CUB~\cite{welinder2010caltech}, Stanford Cars~\cite{krause20133d}, FGVC-Aircraft~\cite{maji2013fine}, CIFAR-100~\cite{krizhevsky2009learning}, Sketches~\cite{eitz2012humans}, WikiArt~\cite{saleh2015large} and Places365~\cite{zhou2017places}) with various ViT variants (including DeiT-Ti~\cite{touvron2021training}, DeiT-S~\cite{touvron2021training} and T2T-ViT-12~~\cite{yuan2021tokens}). Experimental results demonstrate that MEAT exhibits significant superiority to its state-of-the-art CNN counterparts with $4.0\sim 6.0\%$ absolute boosts in accuracy, meanwhile consuming much lower storage cost for saving task-specific masks.
%In conclusion, our contribution is a novel task-incremental learning method tailored for vision transformers with two specifically designed lightweight adaptors for adapting token interaction patterns and trained neurons from the old task to the new tasks. Extensive experiments have been conducted on various image classification benchmarks as new incremental tasks and with popular variants of  vision transformers, \eg, DeiT-Ti~\cite{touvron2021training}, DeiT-S~\cite{touvron2021training}, and T2T-ViT-12~\cite{yuan2021tokens}. Moreover, a large-scale dataset, Plcaces365~\cite{zhou2017places}, is also adopted to validate the effectiveness of our method when adapting initial tasks to large domain shift tasks. Experimental results have demonstrated that our method help vision transformers effectively and efficiently learn new tasks without catastrophic forgetting. Moreover, we further explore and discuss the difference in incremental learning between vision transformers and CNNs.
% In conclusion, we made following three main contributions in this work.

In conclusion, the main contributions of our work are summarized as follows: 
\begin{itemize}
    \item We propose MEAT, the first ViT-backed continual learning method to the best of our knowledge, to advance the development of continual learning with ViTs.
    \item We introduce three innovations into MEAT, including masking partial parameters, avoiding manual hyperparameter setting, and meta-attention mechanism to boost the performance of MEAT.
    \item Extensive experiments demonstrate that MEAT exhibits significant superiority over its state-of-the-art CNN counterparts, meanwhile consuming much lower storage costs for saving task masks.
\end{itemize}

\section{Related Works}
\label{related-works}

\subsection{Vision Transformers}
The transformer~\cite{vaswani2017attention}, a prevailing network architecture in nature language processing~(NLP)~\cite{devlin2018bert, radford2018improving, brown2020gpt3}, has received growing interest and achieved great accomplishment in the computer vision field, enjoying state-of-the-art performance on many visual tasks, including image classification~\cite{dosovitskiy2020image, touvron2021training, liu2021swin}, object detection~\cite{carion2020end, dai2021up}, and object segmentation~\cite{zhang2020feature, chen2021pre}.
% , and graph representation~\cite{ying2021transformers}. 
Among these works, Vision Transformer~\cite{dosovitskiy2020image}, the pioneering work in this area, first introduces a complete transformer-based architecture into image classification tasks by splitting an image into $ 16 \times 16 $ patches and embedding them into a sequence of tokens as the model input like words in NLP. Inspired by the excellent results achieved by Vision Transformer, many researchers have started to study and improve transformer-based models in computer vision. DeiT~\cite{touvron2021training} improves the training efficiency of Vision Transformer by introducing a new distillation token and some training strategies. Swin Transformer~\cite{liu2021swin} presents a new transformer backbone that constructs a hierarchical representation. Tokens-To-Token Vision Transformer~ (T2T-ViT)~\cite{yuan2021tokens} adopts a tokens-to-token (T2T) process to achieve great results trained from scratch on ImageNet~\cite{deng2009imagenet}. 
% To demonstrate the effectiveness of our method, 
% In our approach,
In our experiments
we employ three representative vision transformers: DeiT-Ti, DeiT-S and T2T-ViT-12 as backbone networks. 

% failed for vision transformers
\subsection{Continual Learning}
% Generally, there are two kinds of settings for incremental learning: (1) \textit{task incremental learning} that extends knowledge with new tasks which have clear domain boundaries~\cite{li2017learning, mallya2018piggyback}; (2) \textit{class incremental learning} that accumulates knowledge over different sets of categories separated from the same dataset~\cite{kirkpatrick2017overcoming, rebuffi2017icarl, rusu2016progressive}. In this work, we mainly focus on task incremental learning. To the best of our knowledge, most of the existing studies are designed for convolutional neural networks~(CNNs) in the realm of computer vision.
Continual learning involves incrementally training a model with a new stream of tasks while preserving its previous knowledge basis, which has attracted much interest in recent years~\cite{delange2021continual}.
Generally, there are two kinds of settings for continual learning: (1) \textit{task continual learning} that extends knowledge with new tasks which have clear domain boundaries; (2) \textit{class continual learning} that accumulates knowledge over different sets of categories separated from the same dataset.
In this work, we mainly focus on task continual learning. 
Previous continual learning methods can be broadly categorized into replay methods, regularization methods and mask methods.
The replay methods~\cite{rebuffi2017icarl, chaudhry2019continual, rolnick2018experience, isele2018selective, de2021CoPE} expect to store a subset of samples of previous tasks and retrain the model on old samples to review knowledge of old tasks. The regularization-based methods\cite{li2017learning, jung2016less, kirkpatrick2017overcoming, zenke2017continual, zhang2020class, lee2017imm, paik2020npc} utilize the knowledge distillation technique~\cite{hinton2015distilling},
special regularization terms to avoid catastrophic forgetting.
The other mask methods of continual learning are devoted to expanding the network capacity
via introducing extra masks for each new task~\cite{mallya2018packnet, mallya2018piggyback, masana2021ternary, serra2018overcoming, hung2019compacting, yoon2019scalable, singh2020calibrating}. 
The knowledge of previous tasks can be preserved by sequentially increasing 
new masks~(weight masks~\cite{mallya2018packnet, mallya2018piggyback, masana2021ternary, hung2019compacting, yoon2019scalable} or unit masks~\cite{serra2018overcoming, singh2020calibrating}) and masking out parameters of old tasks simultaneously. For example, PackNet~\cite{mallya2018packnet} iteratively performs pruning a well-trained base network and maintains binary sparsity masks to fix necessary parameters for incoming tasks. Piggyback~\cite{mallya2018piggyback} introduces binary masks on all parameters of a base network for each task without the forgetting problem. HAT~\cite{serra2018overcoming} designs unit masks and keeps the feature embeddings of learned tasks to preserve information of old tasks. Our proposed MEAT builds upon mask methods: given a well-initialized transformer, we assign attention masks to the self-attention mechanism and a portion of parameters to fully leverage the architectural characteristics of ViTs for continual learning.

Besides incremental learning in computer vision mainly designed for CNN structures, a growing body of research in NLP has equipped the transformer with incremental learning. Adaptor-BERT~\cite{houlsby2019parameter} adds two fully connected layers as an adaptor in each layer and freezes old parameters
~(except for normalization layers)
during training. Inspired by this work, B-CL~\cite{ke2021adapting} adopts capsules and dynamic routing~\cite{sabour2017dynamic} to transfer old knowledge to new tasks for aspect sentiment classification tasks.
\cite{huang2021continual} presents an information disentanglement-based regularization method to further generalize old task knowledge. We also compare MEAT with Adaptor-BERT to verify the effectiveness of our methods.

\section{Method}
% \input{floats/fig1.tex}
% In this section, We first briefly revisit the two essential components of transformers, \ie, MHSA and FFN, then
% give the definition of the standard token interaction. Next we dentally describe the proposed MEAT method in two subsection: attention to

% TI adaptor and FFN adaptor are dentally delineated in the following sections, respectively. 
% Furthermore, a penalty term is imposed to avoid extreme sparse interactions between image tokens after applying the proposed TI adaptor. In the end  the relations between our method and existing model-based methods is also discussed.

% In this section, an efficient approach for continually learning new tasks with vision transformers is explored, which is composed of two specially designed adaptors, namely TI adaptor and FFN adaptor, as shown in Figure~\ref{fig:f1}.
% In order to avoid forgetting previously learned tasks, two adaptors are customized for each incoming task via selectively activating or isolating neurons and modify communication between visual tokens based on a well-initialization transformer-based model.
% We first briefly revisit the two essential components of transformers, \ie, MHSA and FFN, then the proposed TI adaptor and FFN adaptor are dentally delineated in the following sections, respectively. 
% Furthermore, a penalty term is imposed to avoid extreme sparse interactions between image tokens after applying the proposed TI adaptor.

\subsection{Preliminaries}
\label{sec3-1}

A typical vision transformer is composed of three key components, \ie, the trainable linear projection for embedding patch features, the multi-head self-attention~(MHSA) block, and the feed-forward network~(FFN) block. An input image is split into $n$ small patches as image tokens, then mapped to a sequence of $d$-dimension vectors. A trainable class token is concatenated to the image token sequence for the final classification. The input sequence $\mathbf{X}\in\mathbb{R}^{(n+1)\times d}$ is fed into a stack of identical encoder layers. Each encoder layer consists of an MHSA block and an FFN block sequentially with residual connections.
Specifically, the MHSA block with $H$ heads can be formulated as
\begin{equation}
    \text{MHSA}(\mathbf{Q}, \mathbf{K}, \mathbf{V}) = \text{Concat}(\text{head}_1, \ldots, \text{head}_H)\mathbf{W}^O \text{,}
    \label{eq:mhsa}
\end{equation}
where $\mathbf{Q}$, $\mathbf{K}$ and $\mathbf{V}$ are the query, key, value embeddings; $\text{head}_h\in\mathbb{R}^{(n+1)\times d_k}$ is the output of attention head $h$ that satisfies $d_k=d/H$, and $\mathbf{W}_h^O\in\mathbb{R}^{d_k \times d}$ is the output projection matrix. The attention head $h$ is calculated by
\begin{equation}
    \text{head}_h = \mathbf{\Psi}_h \mathbf{V}_h = \sigma\left(\mathbf{A}_h\right)\mathbf{V}_h =
    \sigma\left(\frac{\mathbf{Q}_h \mathbf{K}_h^{\top}}{\sqrt{d_k}}\right)\mathbf{V}_h
    \label{eq:attention}
    \textit{,}
\end{equation}
where $\mathbf{Q}_h=\mathbf{X}\mathbf{W}_h^Q$, $\mathbf{K}_h=\mathbf{X}\mathbf{W}_h^K$, and $\mathbf{V}_h=\mathbf{X}\mathbf{W}_h^V$ are linear projections of $\mathbf{X}$ by $\mathbf{W}_h^Q$, $\mathbf{W}_h^K$, and $\mathbf{W}_h^V\in\mathbb{R}^{d\times d_k}$ respectively; $\mathbf{A}_h = \mathbf{Q}_h \mathbf{K}_h^{\top} \ / \sqrt{d_k}$ is the dot-production attention matrix; $\sigma(\cdot)$ is the softmax activation function; $\mathbf{\Psi}_h \in \mathbb{R}^{(n+1)\times(n+1)}$ is an asymmetrical matrix, measuring the similarity between all the pairs of queries and keys by performing dot-production. For instance, the entry $\Psi_h^{i,j}$ of $\mathbf{\Psi}_h$ denotes the attention score that token $i$ pays to token $j$.

% In this paper, we refer to the token interaction pattern like $\Psi_h$, that performs attention computation between all image token pairs~(transformed to quires and keys here) without exception, as the \textit{standard token interaction pattern}. 
% The standard token interaction pattern implies that each image token are equally performed dot product with every token, without exception.
% the  attention  mechanism 
% . In this paper, the attention function that 

The FFN block is composed of two linear layers and an activation function $\phi(\cdot)$~(\eg, GELU~\cite{hendrycks2016gaussian}) and maps the input sequence $\mathbf{X} $ to
\begin{equation}
    \text{FFN}(\textbf{X}; \mathbf{W}_1, \mathbf{W}_2) = \phi(\mathbf{X}\mathbf{W}_1)\mathbf{W}_2 \text{,}
\end{equation}
where $\mathbf{W}_1\in\mathbb{R}^{d\times d'}$, $\mathbf{W}_2\in\mathbb{R}^{d' \times d}$ are projection matrices.
The bias terms of MHSA and FFN are omitted for simplicity.  
\subsection{Meta-Attention}
According to Eqn.~\ref{eq:mhsa} and~\ref{eq:attention}, in the MHSA block, the output of each image token is dependent on all the input tokens. Thus the self-attention mechanism can be generally considered as a dense relationship within all the image token pairs. As a result, all image tokens in the same layer are involved for final classification regardless of the assigned tasks.
In this paper, we refer to the token interaction pattern like $\mathbf{\Psi}_h$, which performs attention computation between all image token pairs equally and densely, as the \textit{standard token interaction pattern}. The proposed MEta-ATtention~(MEAT) aims to dynamically adapt the standard token interaction pattern to the new tasks via putting attention to self-attention.
For simplicity and putting the focus on image token interactions, the class token is kept activated, and $\mathbf{\Psi}_h \in \mathbb{R}^{n \times n}$ only measures the relationship between image tokens.
Furthermore, we also extend the mechanism of MEAT to the trained neurons of the FFN block by paying attention to each neuron, exploring a suitable sub-network of the initial trained weights to boost the continual learning performance. 

\subsubsection{Attention to Self-attention}
\label{sec3-3}

\begin{figure}[!t]
\centering%
    \includegraphics[width=0.98\linewidth]{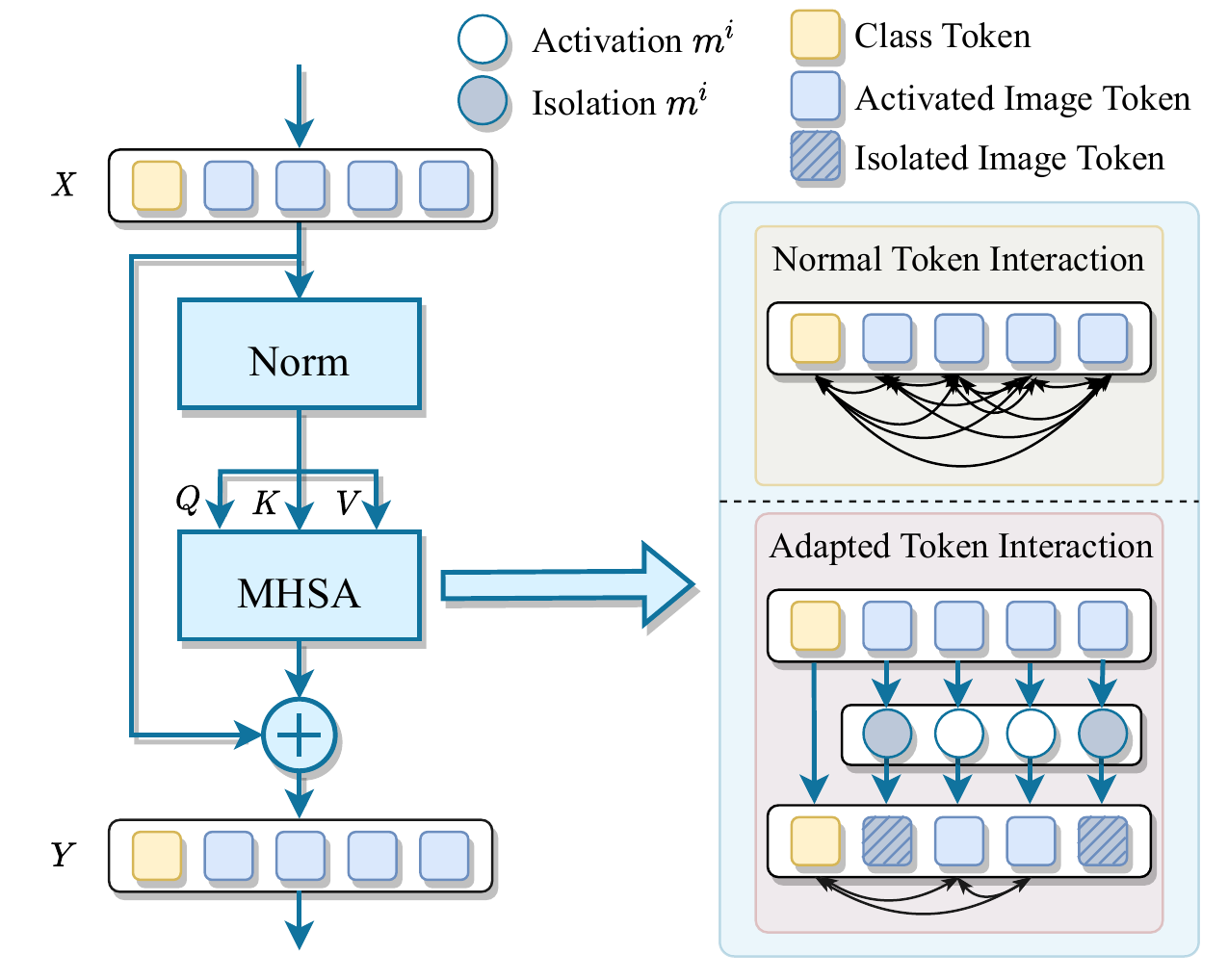}%
    \label{fig:ti-a}%

\caption{Illustration of the working mechanism of MEAT in the MHSA block of a transformer encoder layer. In the standard token integration, all the tokens interact with each other without limit. MEAT proposes attention masks to modify this communication pattern by dynamically activating and isolating image tokens. $X$ and $Y$ denote the input and the output sequence. $Q$, $K$ and, $V$ represent query, key, and value for the MHSA module.}
\label{fig:f1}
\end{figure}

% a novel Token Interaction~(TI) adaptor is 
For a well-initialization transformer as the base model, each image token interacts with each other in the standard information interaction form in the MHSA block for the initial old task. 
MEAT dynamically assigns an attention mask $\mathbf{m} \in \mathbb{R}^{n}$ with continuous values to modify the standard information interaction between all image tokens to learn adaptive communication patterns when sequentially studying new tasks with domain shifts from old tasks. In particular, for the $i$-th input image token, the $i$-th entry of mask $\mathbf{m}$, $m^i$, 
% a continuous variable $t_i$ 
% is assigned as an \textit{meta-attention adaptor}~(MEAT adaptor)  
is used
to modify the attention values related to the token $i$ in this layer. Consequently, the MEAT mask $\mathbf{m}$ generates the adaptive token interaction pattern based on the standard information interaction $\mathbf{\Psi}_h$ in a token-wise manner. 
In the forward propagation, the softmax function $\sigma$ in Eqn.~\ref{eq:attention} can be modified as an adaptive softmax function $\sigma_A$ for calculating task specific attention, \ie, $\mathbf{\Psi}_h = \sigma_A\left(\mathbf{A}_h\right)$. The $i$-th row of similarity $\mathbf{\Psi}_h$ can then be written as
\begin{equation}
\setlength{\abovedisplayskip}{3pt}
    \Psi_h^i =  \begin{bmatrix} \Psi_h^{i,j}
    \end{bmatrix}^n_{j=1} = \begin{bmatrix}
        \frac{m^j \exp  { \left( A_h^{i,j} \right) } }{  \sum\nolimits_{s=1}^n m^s \exp  { \left( A_h^{i,s} \right) } }
    \end{bmatrix}^n_{j=1}  \text{.}
    \label{eq:7}
\setlength{\belowdisplayskip}{3pt}
\end{equation}
Accordingly, $\Psi_h^{i,j}$ represents the modified attention that token $i$ pays to token $j$ via the attention mask $m$. The task-specific relationship provides a task-order invariant solution in modifying token interactions: when inference, each new task only employs the corresponding set of masks and the classifier without interference from other tasks.

% With the increasing number of incoming tasks, the proposed MEAT mask with the 
% % $8$-bit 
% continues values requires much storage memory. 

With the increasing number of incoming tasks, the proposed MEAT mask with continuous values requires a lot of memory space. As shown in Figure~\ref{fig:f1}, a binary value MEAT mask
% that only consists of $1$-bit integers, 
is adopted to replace the former continuous value mask. Specifically, for the token $i$, a binary variable, the attention entry $m^i \in \{0,1\}$ modifies its adapted attention state,
where $1$ and $0$ indicate whether token $i$ is activated or not in the adapted token interaction pattern. With the binarized MEAT mask, $\Psi_h^{i,j}$ in Eqn.~\ref{eq:7} can be computed as
\begin{equation}
    \tilde{\Psi}_h^{i,j} 
    = \begin{cases}
      \frac{  \exp  { \left( A_h^{i,j} \right) }}{\sum\nolimits_{s=1}^n m^s\exp  { \left(  A_h^{i,s} \right) }}  \text{,}
      & \mbox{ if }  m^j = 1 \text{;}\\
        0 \text{,} & \mbox{otherwise} \text{.}
    \end{cases}
\end{equation}
% \begin{equation*}
%     \tilde{\Psi}_h^{i,j} 
%     \!=\! \frac{m^j \exp  { \left( A_h^{i,j} \right) }}{ \sum\limits_{s=1}^n m^s \exp  { \left( A_h^{i,s} \right) }} 
%     \!=\! \begin{cases}
%       \frac{  \exp  { \left( A_h^{i,j} \right) }}{\sum\limits_{s=1}^n \exp  { \left( A_h^{i,s} \right) }}  \text{,}
%       & \mbox{ if }  m^j = 1 \text{;}\\
%         0 \text{,} & \mbox{otherwise} \text{.}
%     \end{cases}
% \end{equation*}

Equipped with the attention masks,
% the binary activated and isolation masks, 
image tokens can be activated dynamically according to new tasks.
However, the binary mask cannot be directly optimized along with the new task classifier when back-propagation, for it belongs to the non-differential categorical distribution, leading to the NP-hard problem~\cite{haastad2001some}. To obtain the binary $m^i$, we hence introduce a differentiable variable parameterized by trainable $t^i \in \mathbb{R}^{2}$, then utilize a novel Gumbel-Softmax estimator~\cite{jang2016categorical} to approximate the discrete Binomial distribution while enabling gradient descent optimization as
% simultaneously as
% during the forward propagation:
\begin{equation}
    m^{i} = \frac{\exp((\log(t^{i,1}) + g^1)/\tau)}{\sum^2_{k=1}\exp((\log(t^{i,k}) + g^k)\tau)},
    \label{eq3}
\end{equation}
where $\tau > 0$ is the temperature, $g^k$ and $g^0$ are sampled from Gumbel distribution $g = -\log(-\log(u))$ with $u \sim \mathrm{Uniform}(0,1)$. The initial weights of $t^{i}$ are independently sampled from distribution $\mathrm{Uniform}(-\gamma, \gamma)$, where $\gamma > 0$ is a hyperparameter. By employing the Gumbel-Softmax trick, the proposed binary mask can be smoothly optimized along with the classifier of the new task via gradient descent.

For inference, updating weights is not required. The relaxation process is replaced with 
\begin{equation}
    m^{i} = \argmax t^{i,j} \text{.}
\end{equation}
In conclusion, MEAT assigns special-designed attention masks to the self-attention block to
create adapted token interaction patterns in a task-specific manner by dynamically activating and isolating corresponding image tokens when incrementally learning new tasks. 
The binary-value mask substantially reduces additional overheads.
The standard token interaction for the initial old task is a special adapted token interaction pattern that the values of each $m^i$ are always set to $1$.

% \subsection{Feed-Forward Network Adaptor}
\subsubsection{Attention To Feed-Forward Network}
\label{sec3-4}
To further boost the performance of task incremental learning, we extend the mechanism of MEAT to the trained neurons of the FFN block by paying attention to each neuron to explore a suitable sub-network of the initial trained weights. An attention mask designed for the FFN block
is also proposed to dynamically activate and isolate neurons in each linear layer in the FFN block. A sub-network of the well-trained FFN block will be generated automatically on the data biases of incoming tasks.
% avoid modifying previously learned neurons, which will lead to catastrophic forgetting. 
% $\mathbf{W} = [w]_{i,j}
Let $\mathbf{W} \in \mathbb{R}^{d_1 \times d_2}$
represent the weight matrix trained on the initial old task of an arbitrary inner-layer in the FFN block, where $d_1$ and $d_2$ are the input and output feature dimensions, respectively.
Note that $\mathbf{W}$ has been optimized on the initial task. Instead of retraining $\mathbf{W}$ for new tasks, we customize an activation map to $\mathbf{W}$ to prevent catastrophic forgetting. Similar to the TI adaptor,
for each neuron $w^{i,j}$ in the weight matrix $W$, the entry of the binary MEAT mask, $m^{i,j} \in \{0,1\}$ stands for its activation state. When introducing a new task, the binary-valued adaptor is adopted to multiply with $w^{i,j}$ as
\begin{equation}
    \tilde{w}^{i,j} = m^{i,j}w^{i,j} = \begin{cases}
        w^{i,j} \textit{,} & \mbox{ if }  m^{i,j} = 1 \text{;}\\
        0 \textit{,} & \mbox{otherwise} \text{.}
    \end{cases}
\end{equation}
With the help of the binary adaptor, the weight can be preserved or discarded according to the new task.
Similar to the optimization procedure of TI adaptors, the Gumbel-Softmax trick is also adopted to tackle the non-differential problem of the attention mask $m$ like Eqn.~\ref{eq3}. 

% FFN adaptors $s_{i,j}$ by sampling from a trainable continuous probability $b_{i}\in\mathbb{R}^2$.

\subsection{Optimization Objective}
% Besides, a new loss function for TI adaptor is designed to makes use of the redundancy existing in image patches and selectively drop tokens on per new-task specific, the absent of patches may cause the accuracy drop when isolated token numbers are beyond normal limits.
The final optimization objective consists of two loss functions. The first one is the conventional cross-entropy loss $\mathcal{L}_{ce}(\hat{p}, p)$, where $\hat{p}$ and $p$ are the predicted category distribution and the ground-truth label.
Isolating image tokens may cause accuracy drops when isolated token numbers are beyond normal limits.
A new drop-control loss is therefore introduced to prevent excessive patch dropping as
\begin{equation}
    \mathcal{L}_{dc}(m) = \frac{1}{L}\sum^L_{l=1}\left(\lambda - \frac{1}{n}\sum^n_{i=1}m^{i}_l \right)^2,
    \label{eq7}
\end{equation}
% \begin{equation}
%     \mathcal{L}_{dc}(m) = \frac{1}{L}\sum^L_{l=1}\left(\frac{\lambda(E-e)}{E} - \frac{1}{n}\sum^n_{i=1}m^{i}_l \right)^2,
%     \label{eq7}
% \end{equation}
% where $e$ and $E$ refer to the current and whole epoch numbers,
where $\lambda$ is a coefficient to adjust the expected activated token numbers. Eqn.~\ref{eq7} regulates the mask by $\lambda$, avoiding isolating too many image tokens at the early training stage, which tends to degrade performance. Let $\alpha$ indicate a weighting factor, the final optimization objective is summed as
\begin{equation}
    \mathcal{L} =  \mathcal{L}_{ce}(\hat{p}, p) + \alpha \mathcal{L}_{dc}(m).
\end{equation}

\section{Experiments}
\label{experiments}
\begin{table*}[!t]
\centering
\resizebox{\textwidth}{!}{
\begin{tabular}{l| l | c| c *{7}{c}}
    \toprule
    % \rowcolor{gray!25}
    & 
    \multicolumn{1}{c|}{\multirow{2}{*}{\textbf{Dataset}}} & 
    \multicolumn{7}{c}{\multirow{1}{*}{\textbf{Method}}} \\
    \hhline{*{2}{|~}*{7}{|-}|~|}
    & & 
    \cellcolor{gray!25} \textbf{Individual}
    & \cellcolor{gray!25} \textbf{Classifier}
    & \cellcolor{gray!25} \textbf{LwF~\cite{li2017learning}}
    & \cellcolor{gray!25} \textbf{Piggyback~\cite{mallya2018piggyback}}
    & \cellcolor{gray!25} \textbf{HAT~\cite{serra2018overcoming}}
    & \cellcolor{gray!25} \textbf{Adaptor-B~\cite{houlsby2019parameter}}
    & \cellcolor{gray!25} \textbf{MEAT} \\

    \toprule

    \multirow{10}{*}{\rotatebox{90}{\textbf{DeiT-Ti}}} 

    & CUB & \small 75.13 & \small 46.05  & \small 59.03 & \small 60.65 & \small \color{blue}68.34 & \small 66.03 & \small \textbf{71.16} \\
    & Cars & \small 69.82 & \small 16.27 & \small 39.39 & \small 44.87 & \small \color{blue}50.57 & \small 45.50 & \small \textbf{53.42} \\ 
    & FGVC & \small 70.00 & \small 14.35 & \small 38.87 & \small 45.58 & \small \color{blue}46.71 & \small 41.28 & \small \textbf{52.69} \\
    & WikiArt & \small 72.13 & \small 38.64 & \small 46.88 &  \small \color{blue}62.42 & \small 61.84 & \small 57.04 & \small \textbf{64.63}  \\
    & Sketches & \small 73.50 & \small 30.64 & \small 53.17 & \small 69.07 & \small 65.49 & \small\color{blue} 69.21 & \small \textbf{70.73} \\
    & CIFAR-100 & \small 83.85 & \small 66.05 & \small 69.79 & \small 71.18 & \small 70.67 & \small \color{blue}75.21 & \small \textbf{78.13} \\
    \cline{2-9}
    &  \multicolumn{1}{c|}{\multirow{2}{*}{
    ImageNet}} & \small 30.82 & \small 72.20 & \small 26.24 & \small 72.20 & \small N/A & \small 72.20 & \small 72.20  \\
    % & (initial task) 
    &  & \small (0.00) & \small (0.00) & \small ({\color{red}$\downarrow 45.96$})
    & \small (0.00) & \small N/A & \small (0.00) & \small (0.00) \\
   \cline{2-9}
    & \multicolumn{1}{c|}{\multirow{2}{*}{
    Model Size}}  & \small 149 MB & \small  23 MB & \small 23 MB & \small 26 MB & \small 23 MB & \small 29 MB & \small 25 MB \\
    
    & & \small ({\color{gray} 6.49x}) & \small ({\color{gray} 0.06x}) &  \small ({\color{gray} 1.00x}) & \small ({\color{gray} 0.21x})& \small ({\color{gray} 1.01x})& \small ({\color{gray} 0.28x}) & \small ({\color{gray} 0.16x}) \\
    % \midrule

    % % \multicolumn{8}{c}{\multirow{1}{*}{
    % % \textbf{DeiT-S}}} \\
    % \midrule
    \toprule
     \multirow{10}{*}{\rotatebox{90}{\textbf{DeiT-S}}} 

    & CUB & \small 82.69 & \small 49.10 & \small 69.34 & \small 72.89 & \small \color{blue}79.67  & \small 77.20 & \small  \textbf{81.53} \\
    & Cars & \small 84.74 & \small 18.29 & \small 74.00 & \small \color{blue}74.72 & \small 73.22 & \small 67.23 & \small \textbf{77.20}  \\
    & FGVC & \small 82.69 & \small 15.51 & \small 55.99 & \small 60.04 & \small \color{blue}62.99  & \small 57.04 & \small \textbf{65.69} \\
    & WikiArt & \small 79.48 & \small 43.85 & \small 65.64 & \small 68.09 & \small 70.43 & \small \color{blue}71.33 & \small \textbf{73.43}  \\
    & Sketches & \small 80.68 & \small 39.80 & \small 70.74 & \small 75.03 & \small \color{blue}74.97 & \small 72.87 & \small \textbf{76.68}  \\
    & CIFAR-100 & \small 89.03 & \small 72.71 & \small 75.67 & \small 79.76 & \small 79.52  & \small \color{blue}84.00 & \small \textbf{85.93} \\
    % \midrule
    \cline{2-9}
    
    & \multicolumn{1}{c|}{\multirow{2}{*}{
    ImageNet}} & \small 49.78 & \small 79.84 & \small 23.01 & \small 79.84 & \small N/A & \small 79.84 & \small 79.84  \\
    % & (initial task)
    &
    &
    \small(0.00)
    & \small (0.00) &
    \small ({\color{red}$\downarrow 56.83$}) &
    \small (0.00)  & \small N/A
    & \small (0.00) & \small (0.00)
    \\    
    % \midrule
    \cline{2-9}
    & \multicolumn{1}{c|}{\multirow{2}{*}{
    Model Size}} & \small 582 MB & \small 86 MB & \small 86 MB  & \small 101 MB  & \small 86 MB & \small 99 MB & \small 96 MB \\
    
    & & \small ({\color{gray} 6.77x}) & \small ({\color{gray} 0.03x}) &  \small ({\color{gray} 1.00x}) & \small ({\color{gray} 0.15x})& \small ({\color{gray} 1.01x})& \small ({\color{gray} 0.17x}) & \small ({\color{gray}0.14x}) \\
    \toprule
    \multirow{10}{*}{\rotatebox{90}{\textbf{T2T-ViT-12}}}

    & CUB & \small 74.47 & \small 26.15 & \small 45.33 & \small 63.57 & \small \color{blue}66.57 & \small 64.31 & \small \textbf{69.90}  \\
    & Cars & \small 72.67 & \small 11.52 & \small 59.01 & \small \color{blue}58.22 & \small 54.63 & \small 53.79 & \small \textbf{61.90} \\
    & FGVC & \small 64.09 & \small 12.46 & \small 42.07 &  \small 51.47 & \small \color{blue}52.69 & \small 48.02 & \small \textbf{53.55}\\
    & WikiArt & \small 73.51 & \small 35.57 & \small 51.24 & \small \color{blue}60.34 & \small 58.53 & \small 59.01 & \small \textbf{61.20}
    \\
    & Sketches & \small 76.60 & \small 18.79 & \small 61.98 & \small 73.07 & \small 71.29 & \small \color{blue}74.02 & \small \textbf{74.75} \\
    & CIFAR-100 & \small 85.03 & \small 33.10 & \small 66.34 & \small 70.98 & \small \color{blue}74.86 & \small 73.58 & \small \textbf{77.42} \\
    
    % \midrule
    \cline{2-9}
    & \multicolumn{1}{c|}{\multirow{2}{*}{
    ImageNet}} & \small 32.62 & \small 55.42 & \small 28.54 & \small 55.42 & \small N/A  & \small 55.42 & \small 55.42  \\
    % & (initial task) 
    &
    &
    \small (0.00)
    & \small (0.00)  &
    \small ({\color{red}$\downarrow 26.88$}) &
    \small (0.00)  & \small N/A
    & \small (0.00) & \small (0.00)
    \\    
    % \midrule
    \cline{2-9}
    & \multicolumn{1}{c|}{\multirow{2}{*}{
    Model Size}} & \small 179 MB & \small 27 MB & \small 27 MB & \small 32 MB & \small 28 MB & \small 36 MB & \small 30 MB\\
    
    & & \small ({\color{gray} 6.63x}) & \small ({\color{gray} 0.07x}) &  \small ({\color{gray} 1.00x}) & \small ({\color{gray} 0.20x})& \small ({\color{gray} 1.02x})& \small ({\color{gray} 0.30x}) & \small ({\color{gray} 0.14x}) \\
    % \midrule

    \bottomrule
\end{tabular}
}
\caption{Comparison of performance on six new tasks added on the initial ImageNet task with three vision transformers. With new tasks sequentially fed into the network, the results are averaged over 6 random orders according to 5 preset seeds. Note that Adaptor-B is the used Adaptor-Bert baseline. Bold fonts and blue fonts represent the best and second-best performance on each new task except Individual~(the ideal setting), respectively. Red values marked with $\downarrow$ in parentheses denote the average performance deterioration on the ImageNet task after continually learning new tasks. Gray values in parentheses refer to the times~($\times$) of retrained model sizes compared to Classifier.}
% , and red \textcolor{red}{-} indicates that the method shows zero-forgetting on ImageNet. introduced by all new tasks
\label{tab:results}
\end{table*}

% This section first describes the experimental settings, followed by the  performance comparison of the main experiments between all involved models. Then we give the visualization and discussion of the MEAT masks. 

% Besides, we also show the effectiveness of adopted components in MEAT and compare transformer-based models with CNN-based models in incremental learning.

% show the effectiveness of each adopted component in our method and analysis of two proposed adaptors.
% % conduct ablation studies to further demonstrate how each adaptor influence the performance on the new task and the old task. 
% Besides, we also compare transformer-based models with CNN-based models in the incremental learning task.

\subsection{Experimental Settings}

% \subsubsection{Datasets}
\textbf{Datasets}~~ImageNet~\cite{deng2009imagenet} is set as the initial task. Six widely-used classification benchmarks are adopted as new visual tasks to be added on the ImageNet initialized vision transformers.
% including CUB~\cite{welinder2010caltech}, Stanford Cars~\cite{krause20133d}, FGVC-Aircraft~\cite{maji2013fine}, Sketches~\cite{eitz2012humans}, WikiArt~\cite{saleh2015large} and  CIFAR-100~\cite{li2017learning}. 
Three fine-grained classification datasets are involved in demonstrating the performance of our method on images with finer granularity than ImageNet, including CUB~\cite{welinder2010caltech}, Stanford Cars~\cite{krause20133d} and FGVC-Aircraft~\cite{maji2013fine}. CIFAR-100~\cite{krizhevsky2009learning}, which has a category hierarchy like Imagenet, is also adopted. Sketches~\cite{eitz2012humans}, WikiArt~\cite{saleh2015large} serve as two art-related datasets which contain pictures of different domains from initial tasks. Besides, we further introduce a large-scale dataset, Places365~\cite{zhou2017places}, to investigate the continually learning ability on large domain shifts data of MEAT. All input images are resized to $224 \times 224$ pixels. 
% The summary statistics of all eight datasets are given in the supplementary material.

% \input{floats/fig-vis-patch.tex}

% \subsubsection{Backbones}
\textbf{Backbones}~~In principle, the specially designed meta-attention mechanism can be applied to any vision transformers. In this paper, three popular transformer-based models are used as the backbone: DeiT-Ti~\cite{touvron2021training}, DeiT-S~\cite{touvron2021training}, T2T-ViT-12~\cite{yuan2021tokens}. We follow the official implementations of adopted three ViTs
and insert our proposed attention masks to self-attention~(only added in the MHSA block) and attention masks to FFN~(only added in the FFN block) in each encoder layer. The official pre-trained weights on ImageNet serve as the well-initialization weights. 
% \footnote{https://github.com/facebookresearch/deit}\footnote{https://github.com/yitu-opensource/T2T-ViT}
% The used training strategy is kept the same as those used in the official code of DeiT. 

% we follow the training strategy used in the official code of DeiT. 

\begin{table}[!t]
\centering
\resizebox{0.48\textwidth}{!}{  
\begin{tabular}{c|c|cccc}
\toprule
\textbf{New Task} & \textbf{Method} & \textbf{DeiT-Ti} & \textbf{DeiT-S} & \textbf{T2T-ViT-12}  \\
\midrule
% \multirow{1}{*}{ImageNet}  & - &  \\
% \midrule
\multirow{3}{*}{Places365} & Individual & 52.67 & 55.22 & 51.53 \\
& Classifier & 40.36 & 44.87 & 37.06 \\
& LwF & 42.17 & 45.13 & 39.77 \\
& Piggyback & 46.32 & 50.71 & 46.38 \\
& MEAT & 48.15 & 52.98 & 47.25\\
\bottomrule
\end{tabular}
}

\caption{Classification results~($\%$) on Places365 dataset, which is added to the ImageNet pretrained transformers.}
\label{tab:results-transformers}
\end{table}

% \subsubsection{\fzl{Parameters.}} 
% \subsubsection{Parameters}
\textbf{Parameters}~~In the training stage, we basically follow the training strategies used in the official code of DeiT~\cite{touvron2021training}. The initial learning rates of new classifiers and the binary MEAT masks are $\frac{batch size}{1024} \times 5e^{-4}$ and $\frac{batch size}{1024} \times 0.1$.
All ViTs are trained for $30$ epochs with a batch size of $256$. 
% $\gamma$,  $\alpha$ and $\beta$ is set to $4$ and weighting factors $\alpha$ and $\beta$ are both $1$.
$\gamma$, $\alpha$ and $\lambda$ are set to $4$, $2$, and $0.9$. All experimental results are averaged over 5 runs on 6 random sequences of new tasks.

% \subsubsection{Compared Baselines}
\textbf{Competitors}~~We compare our proposed MEAT on three transformer-based models with the  following competitors:
(1) \textit{Individual}; (2) \textit{Classifier}; (3) \textit{LwF}~\cite{li2017learning}; (4) \textit{Piggyback}~\cite{mallya2018piggyback}; (5) \textit{HAT}~\cite{serra2018overcoming}; (6) \textit{Adaptor-Bert}~\cite{houlsby2019parameter}. Among these baselines, \textit{Individual} denotes that an independent ViT is trained for each task. As this method duplicates the model by the number of the tasks, its performance can serve as the upper bound of continual learning. \textit{Classifier} is the simple continual learning strategy that finetunes only the classifier layer, \ie, the last fully connected layer, for the new task. LwF, Piggyback, HAT and Adaptor-Bert are four representative existing methods from both computer vision and NLP for continual learing.
%Note that HAT has to store task-specific embeddings. 
For the initial ImageNet task, we directly utilize the official open-source pre-trained weights. For more details, please refer to the supplementary material.

% As a result, the experiments of HAT lack if the embeddings of ImageNet.
%Thus, the performance on ImageNet of HAT has been omitted.
% Given the lack of the embeddings of ImageNet, the performance on ImageNet of HAT has been omitted.

% More details about datasets, hyper-parameters and the training process are described in the supplementary material.
% For more details about datasets and hyper-parameters, please refer to the supplementary material.

% \subsection{\fzl{Comparisions with SOTA Methods}}
\subsection{Benchmark Comparison}

Table~\ref{tab:results} summarizes the main experimental results.
Broadly speaking, our MEAT enjoys superior performance 
with the three ViTs on all the tasks compared to every competitor except \textit{individual}. While \textit{Individual} is ideal from the perspective of accuracy, it increases the model parameters by about 6 times as 6 more independent models are trained for the new tasks, which actually violates the setting of continual learning. Compared with other competitors of continual learning, MEAT adds and retrains only a small number of parameters~(\ie, binary masks). Specifically, \textit{Classifier} does not introduce many extra parameters but shows poor results, especially on data with massive domain shifts from the old task. LwF and HAT require overall retraining of model parameters and suffer from the forgetting problem.
Similar to our proposed method, Piggyback and Adaptor-Bert introduce some additional parameters~(\ie, masks or layers) for continual learning. However, Piggyback applies masks on all parameters. Adaptor-Bert applies two linear layers in each encoder layer leading to excessive extra parameters after its average results. 
In conclusion, our approach balances well in transferring knowledge from the initial task, avoiding catastrophic forgetting, and economizing on parameters. 

% are parameter-based methods using added masks or layers like Ours, 
% % model-expansion approaches like ours, 
% indicating that they can avoid knowledge forgetting and have nothing to do with the sequence of new tasks. However, Piggyback

% couldn't achieve comparable performance like our proposed method, and BERT-adaptor applied two BERT-adaptors in each encoder layer leading to excessive extra parameters after its indifferent results. In conclusion, our approach balances well in transferring knowledge from the initial task, avoiding catastrophic forgetting, and economizing on parameters. 

% \vspace{1em}
% \vspace{-1em}
Another large-scale dataset, Places365~\cite{zhou2017places}, is also introduced as a new task with significant domain shifts from the initial ImageNet task, as shown in Table~\ref{tab:results-transformers}. The experimental results keep nearly the same as those on small datasets. Our proposed MEAT boosts the performance by a large margin, meanwhile averting increasing too many model parameters. 

\subsection{Ablation Study}

\begin{figure}[!t]
\centering
\subfloat[CUB]{\includegraphics[width=0.325\linewidth]{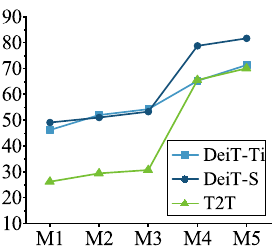}}%
\subfloat[Cars]{\includegraphics[width=0.325\linewidth]{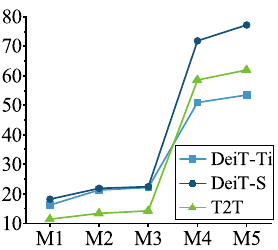}}%
\subfloat[FGVC]{\includegraphics[width=0.325\linewidth]{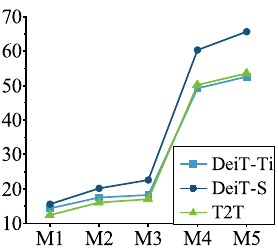}}%
\\
\subfloat[CIFAR-100]{\includegraphics[width=0.325\linewidth]{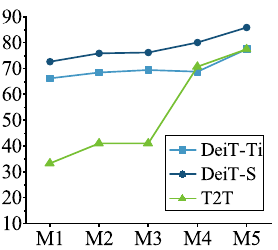}}%
\subfloat[WikiArt]{\includegraphics[width=0.325\linewidth]{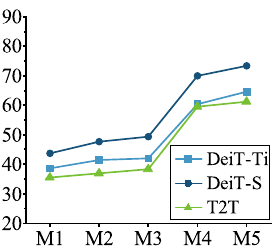}}%
\subfloat[Sketches]{\includegraphics[width=0.325\linewidth]{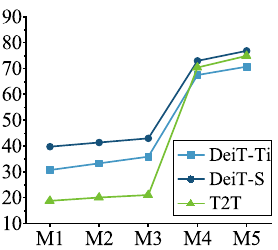}}%

\caption{The effectiveness of each component in our method on six new tasks. In each sub-figure, accuracy~($\%$) of fix model variants
on the same dataset are plotted with three ViTs.}
\label{fig:curv}
\end{figure}

\subsubsection{Effectiveness of Components}
In Figure~\ref{fig:curv}, we demonstrate how our method benefits from each designed component.
Five model variants are listed in each sub-graph to denote three vision transformers equipped with different proposed modules or baselines used for comparison. Concretely, (M1) the Classifier baseline, which is the same as Table~\ref{tab:results}; (M2) transformers with MEAT masks on MHSA, without drop-control loss $\mathcal{L}_{dc}$  in Eqn.~\ref{eq7}; (M3) transformers with MEAT masks on MHSA, with drop-control loss $\mathcal{L}_{dc}$; (M4) transformers with MEAT masks on neurons of the FFN block; (M5) the proposed MEAT. In three vision transformers, both MEAT masks on tokens and neurons, and the loss function efficiently improve the classification accuracy when adding new tasks compared to Classifier baseline~(M1). The MEAT mask on image tokens with the proposed loss function~(M3) achieves $2.67\%\sim7.02\%$ performance boosts without modifying the trained weights, demonstrating the effectiveness of paying attention to the self-attention strategy.
The attention mask on neurons ~(M4) promotes the results considerably by dynamically activating and deactivating pre-trained weights, customizing a sub-network of the complete transformer for each new task. In conclusion, each adopted component of our method consistently promotes continual learning performance.

% which further show the influence of adopted adaptors and optimization objects.
% giving more possibilities. Comparing to the FFN adaptor, the TI adaptor~(M2) brings limited improvements or accuracy degradation sometimes, like DeiT-Ti on FGVC dataset, for isolating excessive image patches leads to information loss in the early training stage. M3 provides a solution to this problem finely. For example, M2 of DeiT-Ti on the CUB dataset causes a performance drop by $0.43\%$, M3 achieves a $1.68\%$ boost instead. Our complete method~(M5) consistently yields sharp performance enhancements to M1,  validating its effectiveness for task incremental learning. In particular, M5 even beats the Fintuning baseline~(M6) in some experiments. For instance, DeiT-S improves $2.19\%$ on CUB, and T2T-S achieves a $3.34\%$ increase on CIFAR-100.

% \input{floats/fig3}

\subsubsection{Comparing to CNNs}

\begin{figure}[!t]
\centering%
\subfloat[Accuracy]{%
    \includegraphics[width=0.48\linewidth]{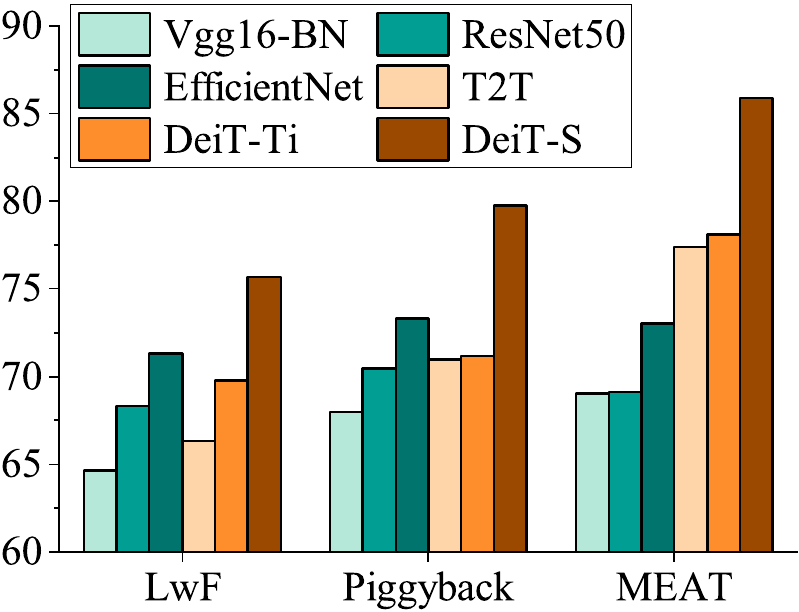}%
    \label{fig:fig3-a}%
}%
\subfloat[Model Size]{%
    \includegraphics[width=0.48\linewidth]{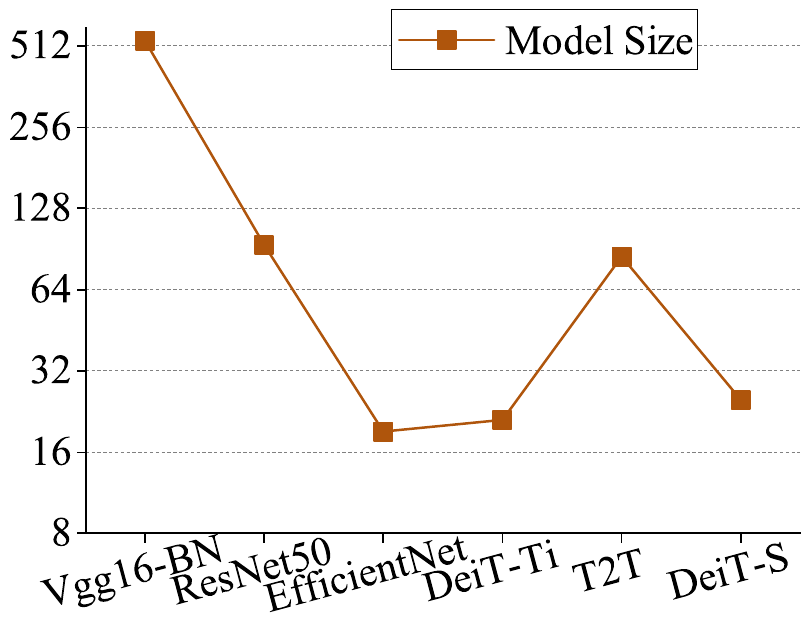}%
    \label{fig:fig3-b}%
}%
\label{fig:fig3}
\caption{(a) Comparing results~($\%$) over vision transformers and CNNs on CIFAR-100. (b) Model size~(MB) comparison.}
\end{figure}

As an emerging model used in computer vision, the ViTs have quite distinct architecture from CNNs. Given that most existing works on incremental learning are based on CNNs, we conduct the experiments on the CIFAR-100 dataset comparing CNNs~(\ie, VGG16-BN~\cite{simonyan2014very}, ResNet50~\cite{he2016deep} and EfficientNet-B4~\cite{tan2019efficientnet}) and vision transformers using LwF, Piggyback, and our proposed MEAT as shown in Figure~\ref{fig:fig3-a}. 
Since the MHSA block only exists in ViTs, in MEAT experiments with CNNs, we only apply the MEAT attention mask to all parameters pre-trained on ImageNet. It can be observed that CNN-based models and transformer-based models show better performance using these three approaches. Nevertheless, due to the absence of the attention mask on self-attention, CNNs under MEAT only have slightly better results than the CNNs under Piggyback. Significantly, the EfficientNet is an excellent network architecture with relatively small model sizes and better performance than DeiT-Ti and T2T in LwF and Piggyback experiments. Nevertheless, using the MEAT masks only achieves similar results with Piggyback. 
In contrast, three ViTs behave much better using MEAT than LwF and Piggyback, because MEAT not only introduces attention masks on a portion of parameters, but also assigns the special attention masks to self-attention and generates unique token interaction patterns for new tasks. 
We attribute this phenomenon to the fundamental idea of our work: paying attention to self-attention.
And the mechanism of self-attention in ViTs builds dense and long-distance dependencies between all image patches. Applied with the MEAT masks to assign different attention values to image tokens, each new task can construct unique token interaction pattern with little overhead.  
However, the masks for CNNs only modifies the activation state of trained neurons, which lacks of modeling the long-distance relationship and brings less performance boosts than MEAT with ViTs. Please refer to the supplementary material for more ablation experimental results and analysis.

\begin{figure*}[!t]%
\centering%
\setlength{\tabcolsep}{0pt}%

\subfloat[DeiT-Ti]{
\begin{tabular}{cc p{0.05cm} cc p{0.05cm} cc p{0.05cm} cc p{0.05cm} cc p{0.05cm} cc}

    \rowcolor{gray!25}

\multicolumn{2}{c}{\textbf{CUB}} & &
\multicolumn{2}{c}{\textbf{Car}} & &
\multicolumn{2}{c}{\textbf{FGVC}} & &
\multicolumn{2}{c}{\textbf{WikiArt}} & &
\multicolumn{2}{c}{\textbf{Sketches}} & &
\multicolumn{2}{c}{\textbf{CIFAR-100}} \\
% \hline
    \rowcolor{gray!25}
layer-2 & layer-11 & & layer-2 & layer-11 & & layer-2 & layer-11 & & layer-2 & layer-11 & & layer-2 & layer-11 & & layer-2 & layer-11 \\
% \hline

\includegraphics[width=1.4cm]{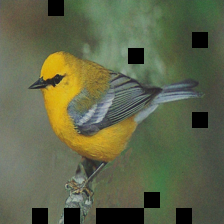} 
&
\includegraphics[width=1.4cm]{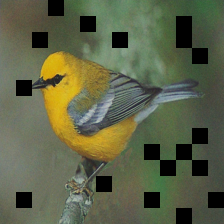} 

&

&
\includegraphics[width=1.4cm]{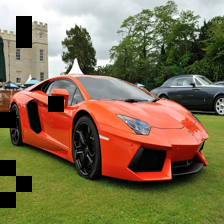} 
&
\includegraphics[width=1.4cm]{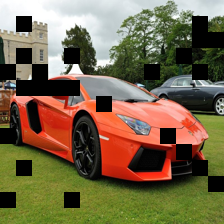} 

&

&
\includegraphics[width=1.4cm]{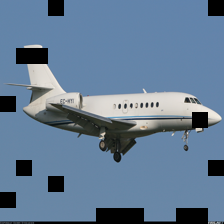} 
&
\includegraphics[width=1.4cm]{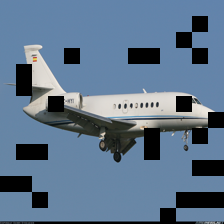}

&

&

\includegraphics[width=1.4cm]{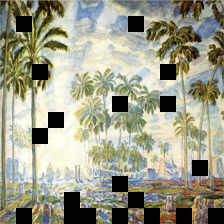} 
&
\includegraphics[width=1.4cm]{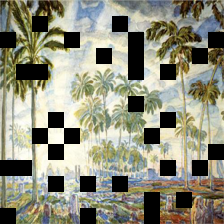}

&

&
\includegraphics[width=1.4cm]{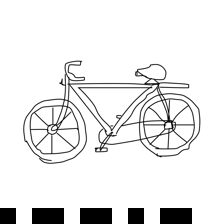} 
&
\includegraphics[width=1.4cm]{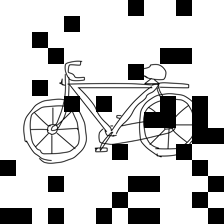}

&

&
\includegraphics[width=1.4cm]{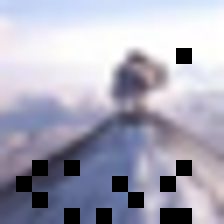} 
&
\includegraphics[width=1.4cm]{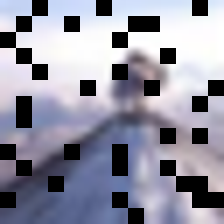} 

\\

\end{tabular}

}%

\subfloat[DeiT-S]{
\begin{tabular}{cc p{0.05cm} cc p{0.05cm} cc p{0.05cm} cc p{0.05cm} cc p{0.05cm} cc}

\includegraphics[width=1.4cm]{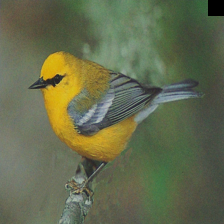} 
&
\includegraphics[width=1.4cm]{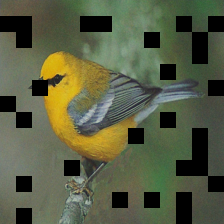} 

&

&
\includegraphics[width=1.4cm]{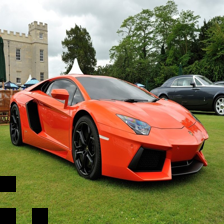} 
&
\includegraphics[width=1.4cm]{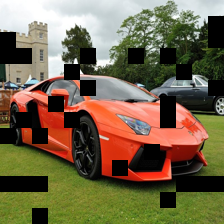} 

&

&
\includegraphics[width=1.4cm]{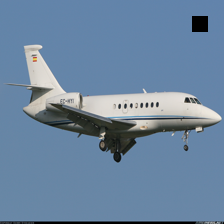} 
&
\includegraphics[width=1.4cm]{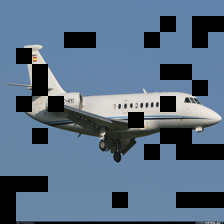}

&

&

\includegraphics[width=1.4cm]{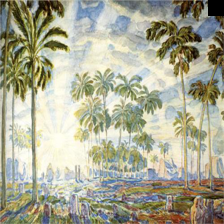} 
&
\includegraphics[width=1.4cm]{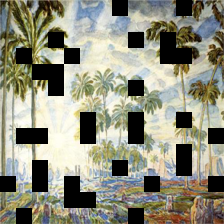}

&

&
\includegraphics[width=1.4cm]{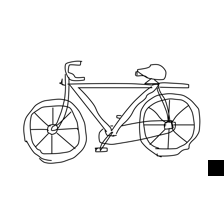} 
&
\includegraphics[width=1.4cm]{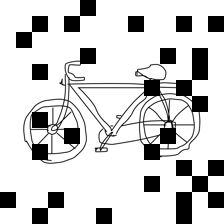}

&

&
\includegraphics[width=1.4cm]{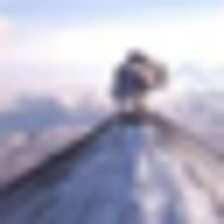} 
&
\includegraphics[width=1.4cm]{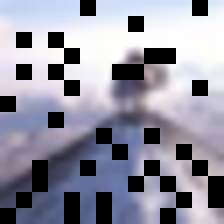} 

\\

\end{tabular}

}%

\subfloat[T2T-ViT-12]{
\begin{tabular}{cc p{0.05cm} cc p{0.05cm} cc p{0.05cm} cc p{0.05cm} cc p{0.05cm} cc}

\includegraphics[width=1.4cm]{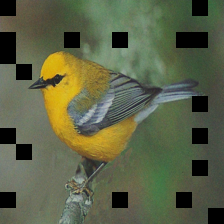} 
&
\includegraphics[width=1.4cm]{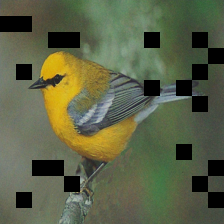} 

&

&
\includegraphics[width=1.4cm]{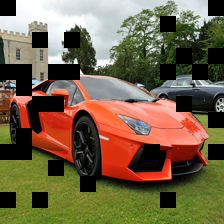} 
&
\includegraphics[width=1.4cm]{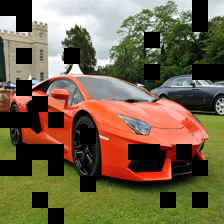} 

&

&
\includegraphics[width=1.4cm]{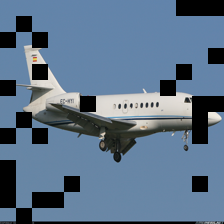} 
&
\includegraphics[width=1.4cm]{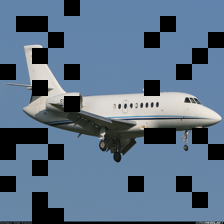}

&

&

\includegraphics[width=1.4cm]{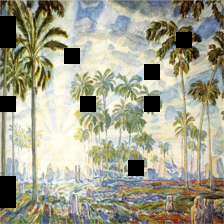} 
&
\includegraphics[width=1.4cm]{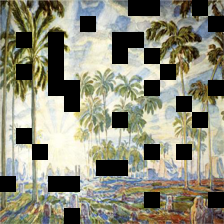}

&

&
\includegraphics[width=1.4cm]{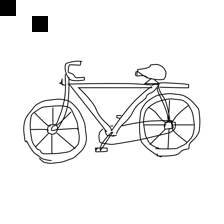} 
&
\includegraphics[width=1.4cm]{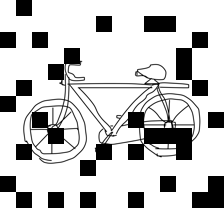}

&

&
\includegraphics[width=1.4cm]{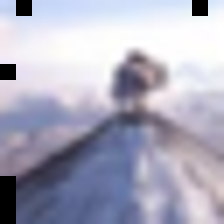} 
&
\includegraphics[width=1.4cm]{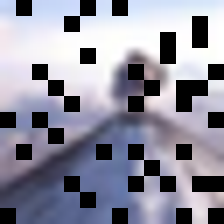} 

\\

\end{tabular}

}%

\caption{Visualization of the trained MEAT masks on the example images of the adopted six datasets at the $2$-th layer and the $11$-th layer of ViTs. The unchanged and black patches are the corresponding locationsfi of the activated and isolated~($1$ and $0$ in masks) image tokens.
}%
% and the rest unchanged image patches are the activated ones.
\label{fig:vis-patch}%
\end{figure*}

\subsection{Analysis and Discussion}

We visualize the trained binary masks of the $2$-th layer and the $11$-th encoder layer in Figure~\ref{fig:vis-patch} on involved six datasets with three adopted ViTs. It can be observed that the activated and isolation states of the same token in the shallow and deep layer present clearly different patterns between datasets and the backbones . First, all the ViTs tend to activate more image tokens at shallow layers and isolate more tokens with the deepening of layers on all new tasks, which is reasonable 
that isolating too many tokens at shallow layers will cause severe information loss. The shallow layer mainly isolates the edge and the background tokens of input images, while the deep layer further isolates more central tokens and focuses on the target object region. For example, on CUB, only some edge tokens are isolated at the shallow layer; at the deep layer more background tokens are dropped and the body tokens of birds are more likely to be activated. 

Another observation is that the trained masks reflect the features of the corresponding datasets. CUB, Car, and FGVC are fine-grained datasets, prone to isolate the background tokens and activate the central tokens containing the target~(\eg, the body of the bird, car, and plane). In contrast, the image of Sketches only contains simple lines and large blank space
and it is prone to isolate more central tokens where the blank space frequently appears. Finally, the bigger model, DeiT-S, tends to retain more tokens than two smaller models at the shallow layer, like on CUB, it only isolates the tokens at the top right-hand corner. This indicates that big models preserve more token interaction at the shallow layers, benefiting the performances on new tasks in Table~\ref{tab:results}.

\begin{figure}[!t]
\centering%
\subfloat[MEAT on tokens]{%
    \includegraphics[scale=0.5]{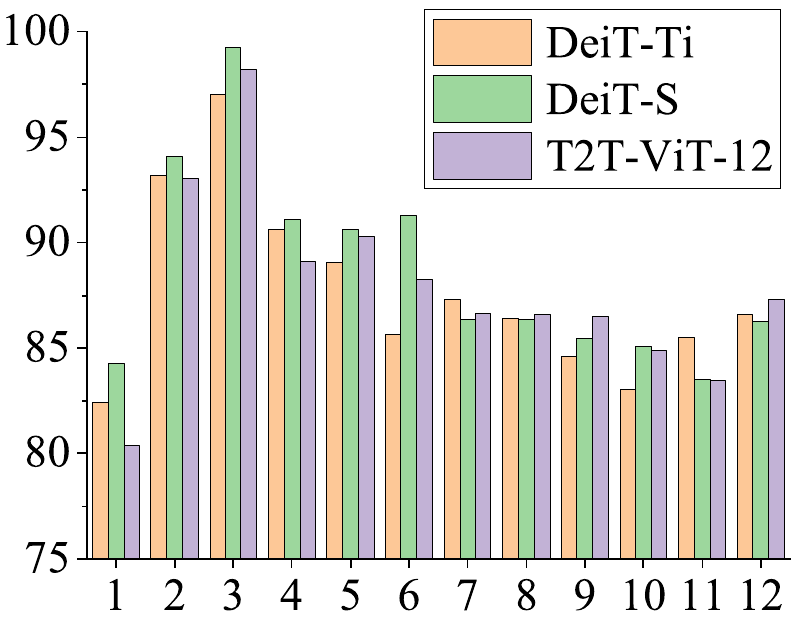}%
    \label{fig:avg-a}%
}%
\subfloat[MEAT on neurons]{%
    \includegraphics[scale=0.5]{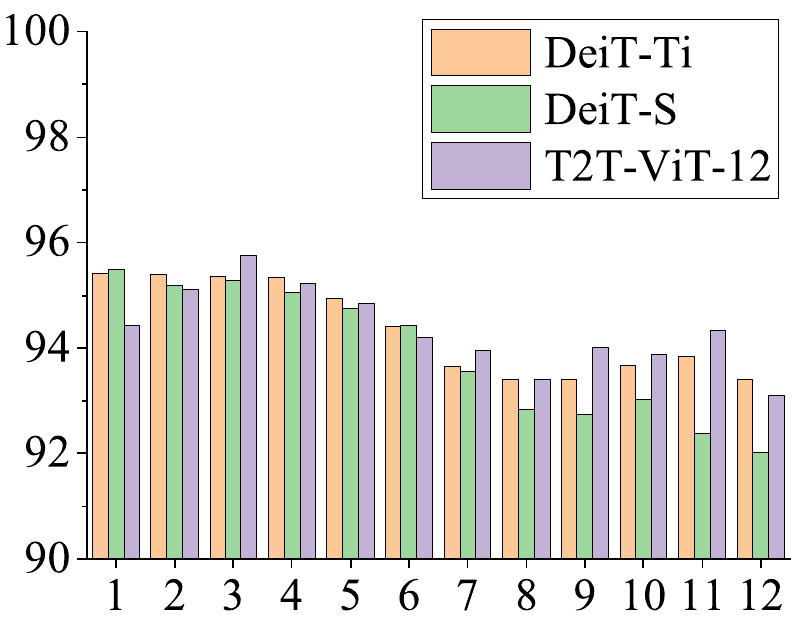}%
    \label{fig:avg-b}%
}%
\caption{The activated rates of (a) image tokens in the MHSA block (b) neurons of the FFN block in each layer averaged on six new tasks of three ViTs using the MEAT attention mask.}
\label{fig:avg}
\end{figure}

We also compare the activated ratios of image tokens and trained neurons averaged on all continual tasks in Figure~\ref{fig:avg}. The attention masks on image tokens in Figure~\ref{fig:avg-a} tend to isolate more tokens at $1$-th layer, activate more tokens at mid-layers, and gradually isolate more tokens at deep layers, which matches with observations of the visualization results. Moreover, it is noticing that the big model, DeiT-S, activates more image tokens than two smaller models at shallow layers, contributing to better results. Figure~\ref{fig:avg-b} gives the activated states of the MEAT masks on FFN neurons. Similar to the token mask, deep layers favor activating fewer neurons than shallow layers. 
% For example, for DeiT-Ti the activating rate of layer $12$ is $93.40\%$ and is $95.39\%$ of layer $2$.
Refer to supplementary material for more visualization results.

% MEAT attention mask. (a) image tokens (a) The average activated token rates on new tasks. (b) The average activated neuron rates of FFN.

% Kept Rates~($\%$) of activated neurons using the FFN adaptor in each transformer layer of three transformers on CIFAR-100. (b) Kept Rates~($\%$) of activated image tokens using the TI adaptor in each layer on CIFAR-100.

\vspace{-1em}
\section{Conclusion and Future Work}
% This paper presents MEAT, a novel task-continual learning method tailored for ViTs named MEAT to pay attention to self-attention for adapting a pre-trained ViT to new tasks.

This paper presents MEAT, a novel task-continual learning method tailored for ViTs, to adapt a pre-trained ViT to new tasks.
MEAT applies attention masks to image tokens in the MHSA block to adaptively generate unique token interaction patterns for new tasks. We further extend the MEAT mechanism to pay attention to neurons for exploring the suitable sub-networks in the FFN block per task. Thus MEAT fully leverages the architecture characteristics of ViTs with task-specific attention masks on self-attention and a portion of parameters without manual hyperparameter setting.
Experiment results show that MEAT effectively improves the performance of continual tasks with little overhead of parameter storage and retraining. 
In future work, we will extend the proposed MEAT beyond the task-continual learning to make further improvements of ViTs in continual learning.

% In future work, we will extend the proposed MEAT beyond the task-continual learning setting to make further improvements of ViTs in continual learning.
% Furthermore, applying replay or regularization methods of continual learning to ViTs is also a promising research topic for future work.
% % Furthermore, applying replay-based or regularization-based continual learning to ViTs is also a promising research topic for future work.

\noindent\textbf{Acknowledgements.}
This work is funded by the National Key R\&D Program of China (Grant No: 2018AAA0101503) and the Science and technology project of SGCC (State Grid Corporation of China): fundamental theory of human-in-the-loop hybrid-augmented intelligence for power grid dispatch and control.

%%%%%%%%% REFERENCES
{\small
\bibliographystyle{ieee_fullname}
\bibliography{reference}

\begin{thebibliography}{10}\itemsep=-1pt

\bibitem{brown2020gpt3}
Tom~B Brown, Benjamin Mann, Nick Ryder, Melanie Subbiah, Jared Kaplan, Prafulla
  Dhariwal, Arvind Neelakantan, Pranav Shyam, Girish Sastry, Amanda Askell,
  et~al.
\newblock Language models are few-shot learners.
\newblock {\em arXiv preprint arXiv:2005.14165}, 2020.

\bibitem{carion2020end}
Nicolas Carion, Francisco Massa, Gabriel Synnaeve, Nicolas Usunier, Alexander
  Kirillov, and Sergey Zagoruyko.
\newblock End-to-end object detection with transformers.
\newblock In {\em ECCV}, pages 213--229. Springer, 2020.

\bibitem{chaudhry2019continual}
Arslan Chaudhry, Marcus Rohrbach, Mohamed Elhoseiny, Thalaiyasingam Ajanthan,
  Puneet~K Dokania, Philip~HS Torr, and M Ranzato.
\newblock Continual learning with tiny episodic memories.
\newblock 2019.

\bibitem{chen2021pre}
Hanting Chen, Yunhe Wang, Tianyu Guo, Chang Xu, Yiping Deng, Zhenhua Liu, Siwei
  Ma, Chunjing Xu, Chao Xu, and Wen Gao.
\newblock Pre-trained image processing transformer.
\newblock In {\em CVPR}, pages 12299--12310, 2021.

\bibitem{dai2021up}
Zhigang Dai, Bolun Cai, Yugeng Lin, and Junying Chen.
\newblock Up-detr: Unsupervised pre-training for object detection with
  transformers.
\newblock In {\em CVPR}, pages 1601--1610, 2021.

\bibitem{de2021CoPE}
Matthias De~Lange and Tinne Tuytelaars.
\newblock Continual prototype evolution: Learning online from non-stationary
  data streams.
\newblock In {\em ICCV}, pages 8250--8259, 2021.

\bibitem{delange2021continual}
Matthias Delange, Rahaf Aljundi, Marc Masana, Sarah Parisot, Xu Jia, Ales
  Leonardis, Greg Slabaugh, and Tinne Tuytelaars.
\newblock A continual learning survey: Defying forgetting in classification
  tasks.
\newblock {\em PAMI}, 2021.

\bibitem{deng2009imagenet}
Jia Deng, Wei Dong, Richard Socher, Li-Jia Li, Kai Li, and Li Fei-Fei.
\newblock Imagenet: A large-scale hierarchical image database.
\newblock In {\em CVPR}, pages 248--255. Ieee, 2009.

\bibitem{devlin2018bert}
Jacob Devlin, Ming-Wei Chang, Kenton Lee, and Kristina Toutanova.
\newblock Bert: Pre-training of deep bidirectional transformers for language
  understanding.
\newblock {\em arXiv preprint arXiv:1810.04805}, 2018.

\bibitem{dosovitskiy2020image}
Alexey Dosovitskiy, Lucas Beyer, Alexander Kolesnikov, Dirk Weissenborn,
  Xiaohua Zhai, Thomas Unterthiner, Mostafa Dehghani, Matthias Minderer, Georg
  Heigold, Sylvain Gelly, et~al.
\newblock An image is worth 16x16 words: Transformers for image recognition at
  scale.
\newblock In {\em ICLR}, 2021.

\bibitem{eitz2012humans}
Mathias Eitz, James Hays, and Marc Alexa.
\newblock How do humans sketch objects?
\newblock {\em TOG}, 31(4):1--10, 2012.

\bibitem{haastad2001some}
Johan H{\aa}stad.
\newblock Some optimal inapproximability results.
\newblock {\em Journal of the ACM (JACM)}, 48(4):798--859, 2001.

\bibitem{he2016deep}
Kaiming He, Xiangyu Zhang, Shaoqing Ren, and Jian Sun.
\newblock Deep residual learning for image recognition.
\newblock In {\em CVPR}, 2016.

\bibitem{hendrycks2016gaussian}
Dan Hendrycks and Kevin Gimpel.
\newblock Gaussian error linear units (gelus).
\newblock {\em arXiv preprint arXiv:1606.08415}, 2016.

\bibitem{hinton2015distilling}
Geoffrey Hinton, Oriol Vinyals, and Jeff Dean.
\newblock Distilling the knowledge in a neural network.
\newblock {\em arXiv preprint arXiv:1503.02531}, 2015.

\bibitem{houlsby2019parameter}
Neil Houlsby, Andrei Giurgiu, Stanislaw Jastrzebski, Bruna Morrone, Quentin
  De~Laroussilhe, Andrea Gesmundo, Mona Attariyan, and Sylvain Gelly.
\newblock Parameter-efficient transfer learning for nlp.
\newblock In {\em ICML}, pages 2790--2799. PMLR, 2019.

\bibitem{huang2021continual}
Yufan Huang, Yanzhe Zhang, Jiaao Chen, Xuezhi Wang, and Diyi Yang.
\newblock Continual learning for text classification with information
  disentanglement based regularization.
\newblock {\em arXiv preprint arXiv:2104.05489}, 2021.

\bibitem{hung2019compacting}
Steven~CY Hung, Cheng-Hao Tu, Cheng-En Wu, Chien-Hung Chen, Yi-Ming Chan, and
  Chu-Song Chen.
\newblock Compacting, picking and growing for unforgetting continual learning.
\newblock {\em arXiv preprint arXiv:1910.06562}, 2019.

\bibitem{isele2018selective}
David Isele and Akansel Cosgun.
\newblock Selective experience replay for lifelong learning.
\newblock In {\em AAAI}, volume~32, 2018.

\bibitem{jang2016categorical}
Eric Jang, Shixiang Gu, and Ben Poole.
\newblock Categorical reparameterization with gumbel-softmax.
\newblock {\em arXiv preprint arXiv:1611.01144}, 2016.

\bibitem{jung2016less}
Heechul Jung, Jeongwoo Ju, Minju Jung, and Junmo Kim.
\newblock Less-forgetting learning in deep neural networks.
\newblock {\em arXiv preprint arXiv:1607.00122}, 2016.

\bibitem{ke2021adapting}
Zixuan Ke, Hu Xu, and Bing Liu.
\newblock Adapting bert for continual learning of a sequence of aspect
  sentiment classification tasks.
\newblock In {\em NAACL}, pages 4746--4755, 2021.

\bibitem{kirkpatrick2017overcoming}
James Kirkpatrick, Razvan Pascanu, Neil Rabinowitz, Joel Veness, Guillaume
  Desjardins, Andrei~A Rusu, Kieran Milan, John Quan, Tiago Ramalho, Agnieszka
  Grabska-Barwinska, et~al.
\newblock Overcoming catastrophic forgetting in neural networks.
\newblock {\em Proceedings of the National Academy of Sciences},
  114(13):3521--3526, 2017.

\bibitem{krause20133d}
Jonathan Krause, Michael Stark, Jia Deng, and Li Fei-Fei.
\newblock 3d object representations for fine-grained categorization.
\newblock In {\em ICCV Workshops}, pages 554--561, 2013.

\bibitem{krizhevsky2009learning}
Alex Krizhevsky, Geoffrey Hinton, et~al.
\newblock Learning multiple layers of features from tiny images.
\newblock 2009.

\bibitem{lee2017imm}
Sang-Woo Lee, Jin-Hwa Kim, Jaehyun Jun, Jung-Woo Ha, and Byoung-Tak Zhang.
\newblock Overcoming catastrophic forgetting by incremental moment matching.
\newblock {\em arXiv preprint arXiv:1703.08475}, 2017.

\bibitem{li2017learning}
Zhizhong Li and Derek Hoiem.
\newblock Learning without forgetting.
\newblock {\em PAMI}, 40(12):2935--2947, 2017.

\bibitem{liu2021swin}
Ze Liu, Yutong Lin, Yue Cao, Han Hu, Yixuan Wei, Zheng Zhang, Stephen Lin, and
  Baining Guo.
\newblock Swin transformer: Hierarchical vision transformer using shifted
  windows.
\newblock {\em ICCV}, 2021.

\bibitem{maji2013fine}
Subhransu Maji, Esa Rahtu, Juho Kannala, Matthew Blaschko, and Andrea Vedaldi.
\newblock Fine-grained visual classification of aircraft.
\newblock {\em arXiv preprint arXiv:1306.5151}, 2013.

\bibitem{mallya2018piggyback}
Arun Mallya, Dillon Davis, and Svetlana Lazebnik.
\newblock Piggyback: Adapting a single network to multiple tasks by learning to
  mask weights.
\newblock In {\em ECCV}, pages 67--82, 2018.

\bibitem{mallya2018packnet}
Arun Mallya and Svetlana Lazebnik.
\newblock Packnet: Adding multiple tasks to a single network by iterative
  pruning.
\newblock In {\em CVPR}, pages 7765--7773, 2018.

\bibitem{masana2021ternary}
Marc Masana, Tinne Tuytelaars, and Joost van~de Weijer.
\newblock Ternary feature masks: zero-forgetting for task-incremental learning.
\newblock In {\em CVPR}, pages 3570--3579, 2021.

\bibitem{mccloskey1989catastrophic}
Michael McCloskey and Neal~J Cohen.
\newblock Catastrophic interference in connectionist networks: The sequential
  learning problem.
\newblock In {\em Psychology of learning and motivation}, volume~24, pages
  109--165. Elsevier, 1989.

\bibitem{paik2020npc}
Inyoung Paik, Sangjun Oh, Taeyeong Kwak, and Injung Kim.
\newblock Overcoming catastrophic forgetting by neuron-level plasticity
  control.
\newblock In {\em AAAI}, volume~34, pages 5339--5346, 2020.

\bibitem{radford2018improving}
Alec Radford, Karthik Narasimhan, Tim Salimans, and Ilya Sutskever.
\newblock Improving language understanding by generative pre-training.
\newblock 2018.

\bibitem{rebuffi2017icarl}
Sylvestre-Alvise Rebuffi, Alexander Kolesnikov, Georg Sperl, and Christoph~H
  Lampert.
\newblock icarl: Incremental classifier and representation learning.
\newblock In {\em CVPR}, pages 2001--2010, 2017.

\bibitem{rolnick2018experience}
David Rolnick, Arun Ahuja, Jonathan Schwarz, Timothy~P Lillicrap, and Greg
  Wayne.
\newblock Experience replay for continual learning.
\newblock {\em arXiv preprint arXiv:1811.11682}, 2018.

\bibitem{rusu2016progressive}
Andrei~A Rusu, Neil~C Rabinowitz, Guillaume Desjardins, Hubert Soyer, James
  Kirkpatrick, Koray Kavukcuoglu, Razvan Pascanu, and Raia Hadsell.
\newblock Progressive neural networks.
\newblock {\em arXiv preprint arXiv:1606.04671}, 2016.

\bibitem{sabour2017dynamic}
Sara Sabour, Nicholas Frosst, and Geoffrey~E Hinton.
\newblock Dynamic routing between capsules.
\newblock {\em arXiv preprint arXiv:1710.09829}, 2017.

\bibitem{saleh2015large}
Babak Saleh and Ahmed Elgammal.
\newblock Large-scale classification of fine-art paintings: Learning the right
  metric on the right feature.
\newblock {\em arXiv preprint arXiv:1505.00855}, 2015.

\bibitem{serra2018overcoming}
Joan Serra, Didac Suris, Marius Miron, and Alexandros Karatzoglou.
\newblock Overcoming catastrophic forgetting with hard attention to the task.
\newblock In {\em ICML}, pages 4548--4557. PMLR, 2018.

\bibitem{simonyan2014very}
Karen Simonyan and Andrew Zisserman.
\newblock Very deep convolutional networks for large-scale image recognition.
\newblock {\em arXiv}, 2014.

\bibitem{singh2020calibrating}
Pravendra Singh, Vinay~Kumar Verma, Pratik Mazumder, Lawrence Carin, and Piyush
  Rai.
\newblock Calibrating cnns for lifelong learning.
\newblock In {\em NeurIPS}, 2020.

\bibitem{tan2019efficientnet}
Mingxing Tan and Quoc Le.
\newblock Efficientnet: Rethinking model scaling for convolutional neural
  networks.
\newblock In {\em ICML}, pages 6105--6114. PMLR, 2019.

\bibitem{touvron2021training}
Hugo Touvron, Matthieu Cord, Matthijs Douze, Francisco Massa, Alexandre
  Sablayrolles, and Herv{\'e} J{\'e}gou.
\newblock Training data-efficient image transformers \& distillation through
  attention.
\newblock In {\em ICML}, pages 10347--10357. PMLR, 2021.

\bibitem{vaswani2017attention}
Ashish Vaswani, Noam Shazeer, Niki Parmar, Jakob Uszkoreit, Llion Jones,
  Aidan~N Gomez, {\L}ukasz Kaiser, and Illia Polosukhin.
\newblock Attention is all you need.
\newblock In {\em NIPS}, pages 5998--6008, 2017.

\bibitem{welinder2010caltech}
Peter Welinder, Steve Branson, Takeshi Mita, Catherine Wah, Florian Schroff,
  Serge Belongie, and Pietro Perona.
\newblock Caltech-ucsd birds 200.
\newblock 2010.

\bibitem{yoon2019scalable}
Jaehong Yoon, Saehoon Kim, Eunho Yang, and Sung~Ju Hwang.
\newblock Scalable and order-robust continual learning with additive parameter
  decomposition.
\newblock {\em arXiv preprint arXiv:1902.09432}, 2019.

\bibitem{yuan2021tokens}
Li Yuan, Yunpeng Chen, Tao Wang, Weihao Yu, Yujun Shi, Zihang Jiang, Francis~EH
  Tay, Jiashi Feng, and Shuicheng Yan.
\newblock Tokens-to-token vit: Training vision transformers from scratch on
  imagenet.
\newblock {\em arXiv preprint arXiv:2101.11986}, 2021.

\bibitem{zenke2017continual}
Friedemann Zenke, Ben Poole, and Surya Ganguli.
\newblock Continual learning through synaptic intelligence.
\newblock In {\em ICML}, pages 3987--3995. PMLR, 2017.

\bibitem{zhang2020feature}
Dong Zhang, Hanwang Zhang, Jinhui Tang, Meng Wang, Xiansheng Hua, and Qianru
  Sun.
\newblock Feature pyramid transformer.
\newblock In {\em ECCV}, pages 323--339. Springer, 2020.

\bibitem{zhang2020class}
Junting Zhang, Jie Zhang, Shalini Ghosh, Dawei Li, Serafettin Tasci, Larry
  Heck, Heming Zhang, and C-C~Jay Kuo.
\newblock Class-incremental learning via deep model consolidation.
\newblock In {\em WACV}, pages 1131--1140, 2020.

\bibitem{zhou2017places}
Bolei Zhou, Agata Lapedriza, Aditya Khosla, Aude Oliva, and Antonio Torralba.
\newblock Places: A 10 million image database for scene recognition.
\newblock {\em PAMI}, 40(6):1452--1464, 2017.

\end{thebibliography}
}

\end{document}